\documentclass[runningheads]{llncs}
\usepackage{graphicx}
\usepackage{comment}
\usepackage{amsmath,amssymb} 
\usepackage{color}
\usepackage{array}
\usepackage{subfigure}
\usepackage{epsfig}
\usepackage{epstopdf}
\usepackage{booktabs}
\usepackage{algorithm}
\usepackage{algpseudocode}
\usepackage{lipsum}
\usepackage[colorlinks,linkcolor=red]{hyperref}

\newcommand{\titlecap}[2]{\textbf{#1} #2}

\begin{document}
\pagestyle{headings}
\mainmatter
\def\ECCVSubNumber{441}  

\newcommand{\eric}[1]{{\color{magenta}[Eric: #1]}}
\newcommand{\yueqi}[1]{{\color{orange}[Yueqi: #1]}}
\newcommand{\he}[1]{{\color{green}[He: #1]}}
\newcommand{\haidong}[1]{{\color{blue}[Haidong: #1]}}
\newcommand{\leo}[1]{{\color{cyan}[Leo: #1]}}
\newcommand{\todo}[1]{{\color{red}[Todo: #1]}}

\title{Curriculum DeepSDF} 

\titlerunning{Curriculum DeepSDF}
%
\author{Yueqi Duan$^{1,*,\dagger}$\qquad
Haidong Zhu$^{2,*}$\qquad 
He Wang$^1$\qquad
Li Yi$^3$ \\
Ram Nevatia$^2$\qquad
Leonidas J. Guibas$^1$}
\authorrunning{Y. Duan, H. Zhu, H. Wang, L. Yi, R. Nevatia and L. J. Guibas}
%
\institute{$^1$Stanford University \ \ 
$^2$University of Southern California\ \
$^3$Google Research}
\maketitle

\newcommand\blfootnote[1]{%
\begingroup 
\renewcommand\thefootnote{}\footnote{#1}%
\addtocounter{footnote}{-1}%
\endgroup 
}

\begin{abstract}
When learning to sketch, beginners start with simple and flexible shapes, and then gradually strive for more complex and accurate ones in the subsequent training sessions. In this paper, we design a ``shape curriculum'' for learning continuous Signed Distance Function (SDF) on shapes, namely \emph{Curriculum DeepSDF}. Inspired by how humans learn, Curriculum DeepSDF organizes the learning task in ascending order of difficulty according to the following two criteria: \emph{surface accuracy} and \emph{sample difficulty}. The former considers stringency in supervising with ground truth, while the latter regards the weights of hard training samples near complex geometry and fine structure. More specifically, Curriculum DeepSDF learns to reconstruct coarse shapes at first, and then gradually increases the accuracy and focuses more on complex local details. Experimental results show that a carefully-designed curriculum leads to significantly better shape reconstructions with the same training data, training epochs and network architecture as DeepSDF. We believe that the application of shape curricula can benefit the training process of a wide variety of 3D shape representation learning methods.\blfootnote{* Equal contribution \ \ \ \ $\dagger$ Corresponding author: duanyq19@stanford.edu}  

\end{abstract}

\section{Introduction}
In recent years, 3D shape representation learning has aroused much attention \cite{qi2017pointnet,qi2017pointnet++,maturana2015voxnet,groueix2018papier,park2019deepsdf}. Compared with images indexed by regular 2D grids, there has not been a single standard representation for 3D shapes in the literature. Existing 3D shape representations can be cast into several categories including: point-based~\cite{qi2017pointnet,qi2017pointnet++,wang2019dynamic,duan2019structural,achlioptas2018learning,rao2020global}, voxel-based~\cite{wu20153d,maturana2015voxnet,qi2016volumetric,choy20163d}, mesh-based~\cite{groueix2018papier,guo20153d,wang2018pixel2mesh,sinha2016deep}, and multi-view~\cite{su2015multi,qi2016volumetric,tulsiani2017multi}.

More recently, implicit function representations have gained an increasing amount of interest due to their high fidelity and efficiency. An implicit function depicts a shape through assigning a gauge value to each point in the object space~\cite{park2019deepsdf,mescheder2019occupancy,chen2019learning,michalkiewicz2019deep}. Typically, a negative, a positive or a zero gauge value represents that the corresponding point lies inside, outside or on the surface of the 3D shape. Hence, the shape is implicitly encoded by the iso-surface (e.g., zero-level-set) of the function, which can then be rendered by Marching Cubes~\cite{lorensen1987marching} or similar methods. Implicit functions can also be considered as a shape-conditioned binary classifier whose decision boundary is the surface of the 3D shape. As each shape is represented by a continuous field, it can be evaluated at arbitrary resolution, irrespective of the resolution of the training data and limitations in the memory footprint.

\begin{figure}[tb]
\centering
\includegraphics[width=0.88\textwidth]{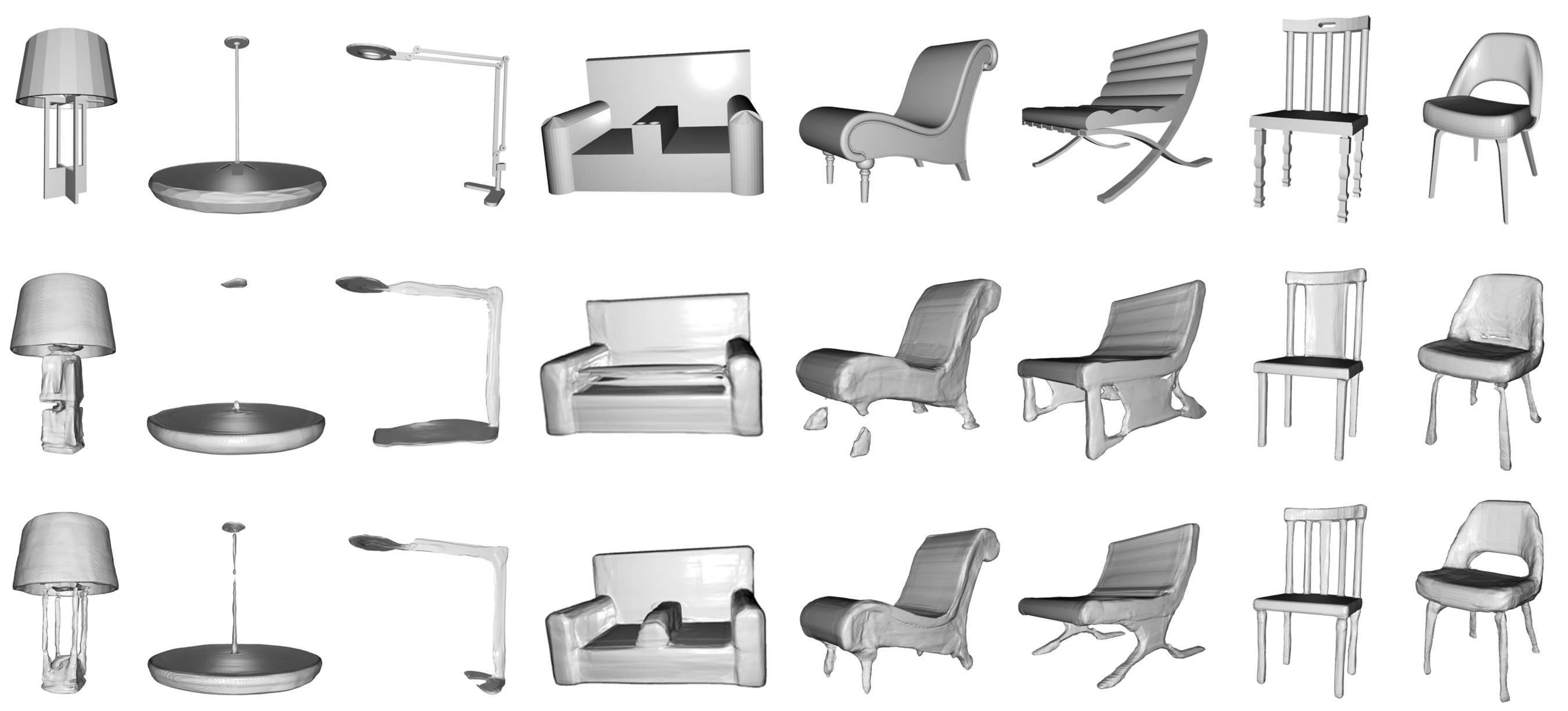}
\caption{3D reconstruction results of shapes with complex local details. From top to bottom: ground truth, DeepSDF~\cite{park2019deepsdf}, and Curriculum DeepSDF. We observe that the network benefits from the designed shape curriculum so as to better reconstruct local details. It is worth noting that the training data, training epochs and network architecture are the same for both methods.}
\label{fig:vis}
\end{figure}

One of the main challenges in implicit function learning lies in accurate reconstruction of shape surfaces, especially around complex or fine structure. Fig.~\ref{fig:vis} shows some 3D shape reconstruction results where we can observe that DeepSDF~\cite{park2019deepsdf} fails to precisely reconstruct complex local details. Note that the implicit function is less smooth in these areas and hence difficult for the network to parameterize precisely. Furthermore, as the magnitudes of SDF values inside small parts are usually close to zero, a tiny mistake may lead to a wrong sign, resulting in inaccurate surface reconstruction. 

Inspired by the works on curriculum learning~\cite{elman1993learning,bengio2009curriculum}, we aim to address this problem in learning SDF by \emph{starting small}: starting from easier geometry and gradually increasing the difficulty of learning. In this paper, we propose a Curriculum DeepSDF method for shape representation learning. We design a shape curriculum where we first teach the network using coarse shapes, and gradually move on to more complex geometry and fine structure once the network becomes more experienced. In particular, our shape curriculum is designed according to two criteria: \emph{surface accuracy} and \emph{sample difficulty}. We consider these two criteria both important and complementary to each other for shape representation learning: \emph{surface accuracy} cares about the stringency in supervising with training loss, while \emph{sample difficulty} focuses on the weights of hard training samples containing complex geometry.

\textbf{Surface accuracy.} We design a tolerance parameter $\varepsilon$ that allows small errors in estimating the surfaces. Starting with a relatively large $\varepsilon$, the network aims for a smooth approximation, focusing on the global structure of the target shape and ignoring hard local details. Then, we gradually decrease $\varepsilon$ to expose more shape details until $\varepsilon=0$. We also use a shallow network to reconstruct coarse shapes at the beginning and then progressively add more layers to learn more accurate details.


\textbf{Sample difficulty.} Signs greatly matter in implicit function learning. The points with incorrect sign estimations lead to significant errors in shape reconstruction, suggesting that we treat these as hard samples during training. We gradually increase the weights of hard and semi-hard\footnote{Here, semi-hard samples are with the correct sign estimations but close to the boundary. In practice, we also decrease the weights of easy samples to avoid overshooting.} training samples to make the network more and more focused on difficult local details.

One advantage of curriculum shape representation learning is that, it provides a training path for the network to start from coarse shapes and finally reach fine-grained geometries. At the beginning, it is substantially more stable for the network to reconstruct coarse surfaces with the complex details omitted. Then, we continuously ask for more accurate shapes which are relatively simple tasks, benefiting from the previous reconstruction results. Lastly, we focus on hard samples to obtain complete reconstruction with precise shape details. This training process can help avoid poor local minima as compared with learning to reconstruct the precise complex shapes directly. Fig.~\ref{fig:vis} shows that Curriculum DeepSDF obtains better reconstruction accuracy than DeepSDF. Experimental results illustrate the effectiveness of the designed shape curriculum. Code will be available at \url{https://github.com/haidongz-usc/Curriculum-DeepSDF}.

In summary, the key contributions of this work are:
\begin{enumerate}
    \item We design a shape curriculum for shape representation learning, starting from coarse shapes to complex details. The curriculum includes two aspects of \emph{surface accuracy} and \emph{sample difficulty}.
    
    \item For surface accuracy, we introduce a tolerance parameter $\varepsilon$ in the training objective to control the smoothness of the learned surfaces. We also progressively grow the network according to different training stages.
    
    \item For sample difficulty, we define hard, semi-hard and easy training samples for SDF learning based on sign estimations. We re-weight the samples to make the network gradually focus more on hard local details.
    
\end{enumerate}

\section{Related Work}

\subsubsection{Implicit function.} Different from point-based, voxel-based, mesh-based and multi-view methods which explicitly represent shape surfaces, implicit functions aim to learn a continuous field and represent the shape with the iso-surface. Conventional implicit function based methods include~\cite{carr2001reconstruction,shen2004interpolating,turk1999shape,turk2002modelling,ohtake2005multi}. For example, Carr~\emph{et al.}~\cite{carr2001reconstruction} used polyharmonic Radial Basis Functions (RBFs) to implicitly model the surfaces from point clouds. Shen~\emph{et al.}~\cite{shen2004interpolating} created implicit surfaces by moving least squares. In recent years, several deep learning based methods have been proposed to capture more complex topologies~\cite{park2019deepsdf,mescheder2019occupancy,chen2019learning,genova2019learning,saito2019pifu,liu2019learning,michalkiewicz2019deep,liao2018deep,xu2019disn,genova2019deep,gropp2020implicit}. For example, Park~\emph{et al.}~\cite{park2019deepsdf} proposed DeepSDF by learning an implicit field where the magnitude represents the distance to the surface and the sign shows whether the point lies inside or outside of the shape. Mescheder~\emph{et al.}~\cite{mescheder2019occupancy} presented Occupancy Networks by approximating the 3D continuous occupancy function of the shape, which indicates the occupancy probability of each point. Chen and Zhang~\cite{chen2019learning} proposed IM-NET by only encoding the signs of SDF, which can be used for representation learning (IM-AE) and shape generation (IM-GAN). Saito~\emph{et al.}~\cite{saito2019pifu} and Liu~\emph{et al.}~\cite{liu2019learning} learned implicit surfaces of 3D shapes from 2D images. These methods show promising results in 3D shape representation. However, the challenges still remains to reconstruct the local details accurately. Instead of proposing new implicit functions, our approach studies how to design a curriculum of shapes for more effective model training.

\subsubsection{Curriculum learning.} The idea of curriculum learning can be at least traced back to~\cite{elman1993learning}. Inspired by the learning system of humans, Elman~\cite{elman1993learning} demonstrated the importance of starting small in neural network training. Sanger~\cite{sanger1994neural} extended the idea to robotics by gradually increasing the difficulty of the task. Bengio~\emph{et al.}~\cite{bengio2009curriculum} further formalized this training strategy and explored curriculum learning in various cases including vision and language tasks. They introduced one formulation of curriculum learning by using a family of functions $L_\mu (\theta)$, where $L_0$ is the highly smoothed version and $L_1$ is the real objective. One could start with $L_0$ and gradually increase $\mu$ to 1, keeping $\theta$ at a local minimum of $L_\mu (\theta)$. They also explained the advantage of curriculum learning as a continuation method~\cite{allgower2003introduction}, which could benefit the optimization of a non-convex training criterion to find better local minima. Graves~\emph{et al.}~\cite{graves2017automated} designed an automatic curriculum learning method by automatically selecting the training path to address the sensitivity of progression mode. Recently, curriculum learning has been successfully applied to varying tasks~\cite{schroff2015facenet,duan2019deep,ilg2017flownet,karras2017progressive,sharma2018improved,jiang2018mentornet,weinshall2018curriculum,hacohen2019power}. For example, deep metric learning methods learn hierarchical mappings by gradually selecting hard training samples~\cite{lu2017discriminative,duan2020deep,sun2020circle}. FaceNet~\cite{schroff2015facenet} proposed an online negative sample mining strategy for face recognition, which was improved by DE-DSP~\cite{duan2019deep} to learn a discriminative sampling policy. 
Progressive growing of GANs~\cite{karras2017progressive,sharma2018improved} learned to sequentially generate images from low-resolution to high-resolution, and also grew both generator and discriminator symmetrically. 
Although curriculum learning has improved the performance of many tasks, the problem of how to design a curriculum for 3D shape representation learning still remains. Unlike 2D images where the pixels are regularly arranged, 3D shapes usually have irregular structures, which makes the effective curriculum design more challenging. 

\section{Proposed Approach}
Our shape curriculum is designed based on DeepSDF~\cite{park2019deepsdf}, which is a popular implicit function based 3D shape representation learning method. In this section, we first review DeepSDF and then describe the proposed Curriculum DeepSDF approach. Finally, we introduce the implementation details.

\subsection{Review of DeepSDF~\cite{park2019deepsdf}}
DeepSDF is trained on a set of $N$ shapes $\{X_i\}$, where $K$ points $\{x_j\}$ are sampled around each shape $X_i$ with the corresponding SDF values $\{s_j\}$ precomputed. This results in $K$ (point, SDF value) pairs:
\begin{eqnarray} \label{sdf}
X_i := \{(x_j,s_j) : s_j=SDF^i(x_j)\},
\end{eqnarray}
A deep neural network $f_\theta(z_i,x)$ is trained to approximate SDF values of points $x$, with an input latent code $z_i$ representing the target shape.

The loss function given $z_i$, $x_j$ and $s_j$ is defined by the $L_1$-norm between the estimated and ground truth SDF values:
\begin{eqnarray} \label{opt_L}
L(f_\theta(z_i,x_j),s_j) = |\text{clamp}_{\delta}(f_\theta(z_i,x_j)) - \text{clamp}_{\delta}(s_j)|,
\end{eqnarray}
where $\text{clamp}_{\delta}(s) := \text{min}(\delta,\text{max}(-\delta, s))$ uses a parameter $\delta$ to clamp an input value $s$. For simplicity, we use $\bar{s}$ to represent a clamping function with $\delta=0.1$ in the rest of the paper.

DeepSDF also designs an auto-decoder structure to directly pair a latent code $z_i$ with a target shape $X_i$ without an encoder. Please refer to~\cite{park2019deepsdf} for more details. At training time, $z_i$ is randomly initialized from $\mathcal{N}(0,0.01^2)$ and optimized along with the parameters $\theta$ of the network through back-propagation:
\begin{eqnarray}
\arg\min_{\theta,z_i}\sum_{i=1}^N\left(\sum_{j=1}^K L(f_\theta(z_i,x_j),s_j) + \frac{1}{\sigma^2}||z_i||_2^2\right),
\end{eqnarray}
where $\sigma = 10^{-2}$ is the regularization parameter.

At inference time, an optimal $z$ can be estimated with the network fixed:
\begin{eqnarray}
\hat{z} = \arg\min_{z}\sum_{j=1}^K L(f_\theta(z,x_j),s_j) + \frac{1}{\sigma^2}||z||_2^2.
\end{eqnarray}





\subsection{Curriculum DeepSDF}
Different from DeepSDF which trains the network with a fixed objective all the time, Curriculum SDF starts from learning smooth shape approximations and then gradually strives for more local details. We carefully design the curriculum from the following two aspects: \emph{surface accuracy} and \emph{sample difficulty}.

\subsubsection{Surface accuracy.} A smoothed approximation for a target shape could capture the global shape structure without focusing too much on local details, and thus is a good starting point for the network to learn. With a changing smoothness level at different training stages, more and more local details can be exposed to improve the network. Such smoothed approximations could be generated by traditional geometry processing algorithms. However, the generation process is time-consuming, and it is also not clear whether such fixed algorithmic routines could meet the needs of network training. In this paper, we address the problem from another view by introducing surface error tolerance $\varepsilon$ which represents the upper bound of the allowed errors in the predicted SDF values.
We observe that starting with relatively high surface error tolerance, the network tends to omit complex details and aims for a smooth shape approximation. Then, we gradually reduce the tolerance to expose more details.

\begin{figure}[tb]
\centering
\includegraphics[width=0.64\textwidth]{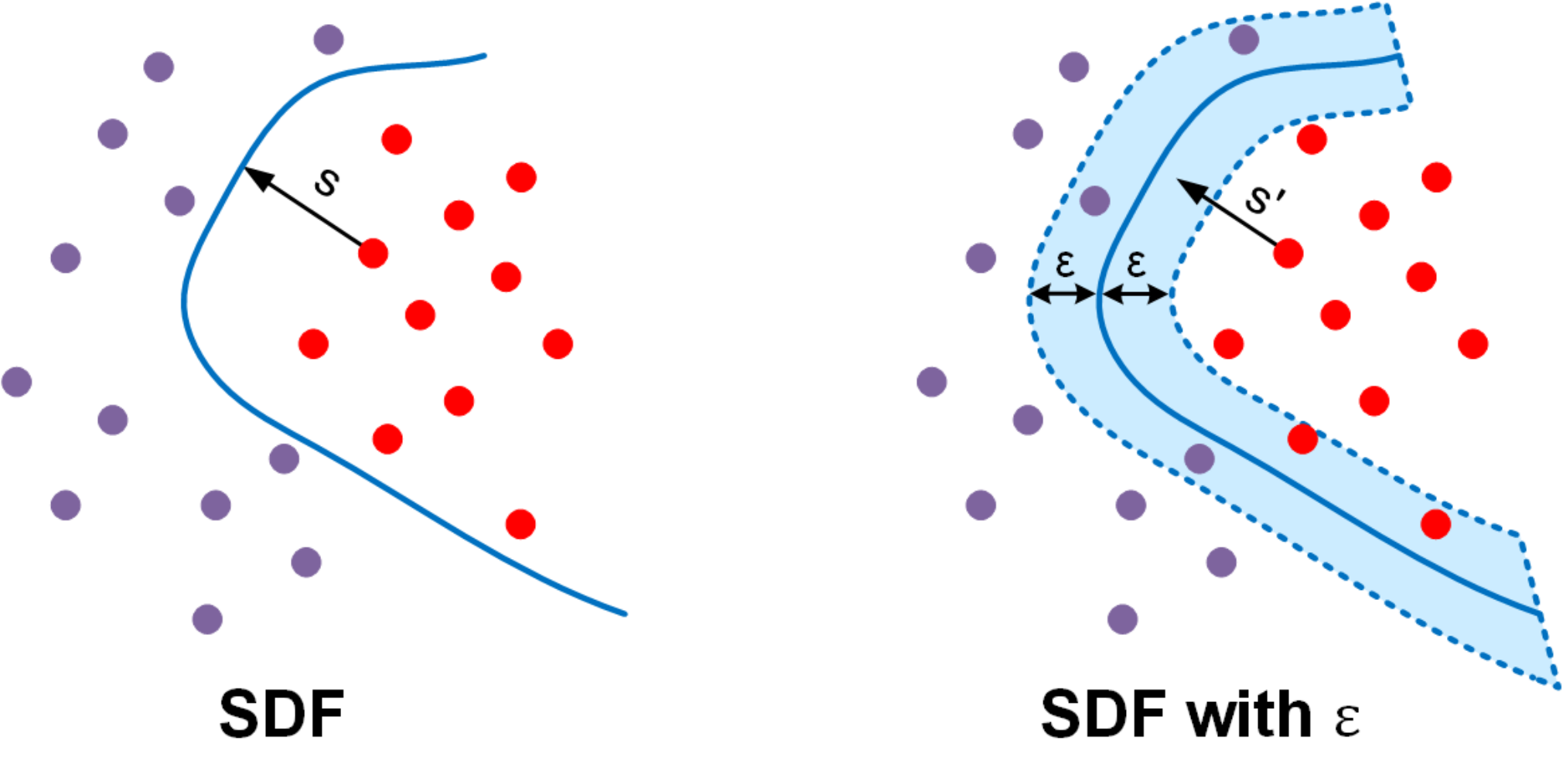}
\caption{The comparison between original SDF and SDF with the tolerance parameter $\varepsilon$. With the tolerance parameter $\varepsilon$, all the surfaces inside the tolerance zone are considered correct. The training of Curriculum DeepSDF starts with a relative large $\varepsilon$ and then gradually reduces it until $\varepsilon=0$.}
\label{fig:tol}
\end{figure}

More specifically, we allow small mistakes for the SDF estimation within the range of $[-\varepsilon, \varepsilon]$ for Curriculum DeepSDF. In other words, all the estimated SDF values whose errors are smaller than $\varepsilon$ are considered correct without any punishment, and we can control the difficulty of the task by changing $\varepsilon$. Fig.~\ref{fig:tol} illustrates the physical meaning of the tolerance parameter $\varepsilon$. Compared with DeepSDF which aims to reconstruct the exact surface of the shape, Curriculum DeepSDF provides a tolerance zone with the thickness of $2\varepsilon$, and the objective becomes to reconstruct any surface in the zone. At the beginning of network training, we set a relatively large $\varepsilon$ which allows the network to learn general and smooth surfaces in a wide tolerance zone. Then, we gradually decrease $\varepsilon$ to expose more details and finally set $\varepsilon=0$ to predict the exact surface.

We can formulate the objective function with $\varepsilon$ as follows:
\begin{eqnarray} \label{opt_Leps}
L_\varepsilon(f_\theta(z_i,x_j),s_j) = \max\{|\bar{f}_\theta(z_i,x_j) - \bar{s}_j|-\varepsilon, 0\},
\end{eqnarray}
where (\ref{opt_Leps}) will degenerate to (\ref{opt_L}) if $\varepsilon=0$.

Unlike most recent curriculum learning methods that rank training samples by difficulty~\cite{weinshall2018curriculum,hacohen2019power}, our designed curriculum on shape accuracy directly modifies the training loss. It follows the formulation in~\cite{bengio2009curriculum} and also has a clear physical meaning for the task of SDF estimation. It is also relevant to label smoothing methods, where our curriculum has clear geometric meanings by gradually learning more precise shapes. We summarize the two advantages of the tolerance parameter based shape curriculum as follows:
\begin{enumerate}
    \item We only need to change the hyperparameter $\varepsilon$ to control the surface accuracy, instead of manually creating series of smooth shapes. The network automatically finds the surface that is easy to learn in the tolerance zone.
    
    \item For any $\varepsilon$, the ground truth surface of the original shape is always an optimal solution of the objective, which has good optimization consistency.
\end{enumerate}

\begin{figure}[tb]
\centering
\includegraphics[width=\textwidth]{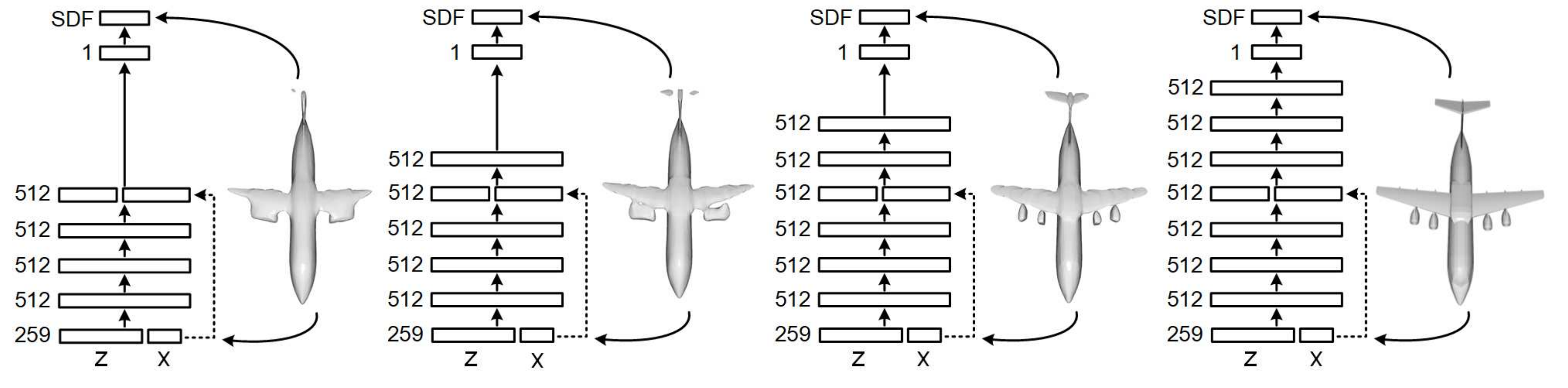}
\caption{The network architecture of Curriculum DeepSDF. We apply the same final network architecture with DeepSDF for fair comparisons, which contains 8 fully connected layers followed by hyperbolic tangent non-linear activation to obtain SDF value. The input is the concatenation of latent vector $z$ and 3D point $x$, which is also concatenated to the output of the fourth layer. When $\varepsilon$ decreases during training, we add one more layer to learn more precise shape surface.}
\label{fig:net}
\end{figure}

In addition to controlling the surface accuracy by the tolerance parameter, we also use a shallow network to learn coarse shapes with a large $\varepsilon$, and gradually add more layers to improve the surface accuracy when $\varepsilon$ decreases. This idea is mainly inspired by~\cite{karras2017progressive}. Fig.~\ref{fig:net} shows the network architecture of the proposed Curriculum DeepSDF, where we employ the same network as DeepSDF for fair comparisons. After adding a new layer with random initialization to the network, the well-trained lower layers may suffer from sudden shocks if we directly train the new network in an end-to-end manner. Inspired by~\cite{karras2017progressive}, we treat the new layer as a residual block with a weight of $\alpha$, where the original link has a weight of $1-\alpha$. We linearly increase $\alpha$ from 0 to 1, so that the new layer can be faded in the original network smoothly. 

\subsubsection{Sample difficulty.} In DeepSDF, the sampled points $\{x_j\}$ in $X_i$ all share the same weights in training, which presumes that every point is equally important. However, this assumption may result in the following two problems for reconstructing complex local details:
\begin{enumerate}
    \item Points depicting local details are usually undersampled, and they could be ignored by the network during training due to their small population. We take the second lamp in Fig.~\ref{fig:vis} as an example. The number of sampled points around the lamp rope is nearly 1/100 of all the sampled points, which is too small to affect the network training.
    
    \item In these areas, the magnitudes of SDF values are small as the points are close to surfaces (e.g. points inside the lamp rope). Without careful emphasis, the network could easily predict the wrong signs. Followed by a surface reconstruction method like Marching Cubes, the wrong sign estimations will further lead to inaccurate surface reconstructions.
\end{enumerate}

To address these issues, we weight the sampled points differently during training. An intuitive idea is to locate all the complex local parts at first, and then weight or sort the training samples according to some difficulty measurement~\cite{bengio2009curriculum,weinshall2018curriculum,hacohen2019power}. However, it is difficult to detect complex regions and rank the difficulty of points exactly. In this paper, we propose an adaptive difficulty measurement based upon the SDF estimation of each sample and re-weight the samples to gradually emphasize more on hard and semi-hard samples on the fly.

\begin{figure}[tb]
\centering
\subfigure[$s>0$]{
\includegraphics[width=0.45\textwidth]{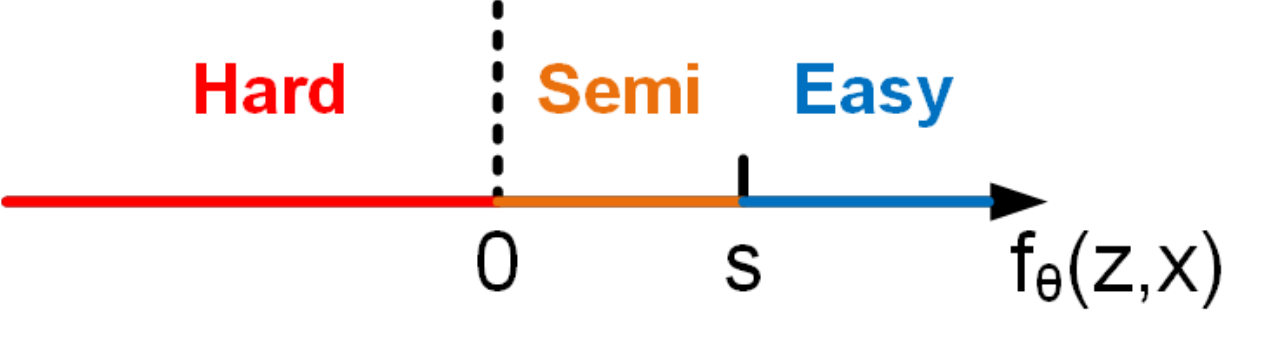}}
	\hspace{0.05in}
\subfigure[$s<0$]{
\includegraphics[width=0.45\textwidth]{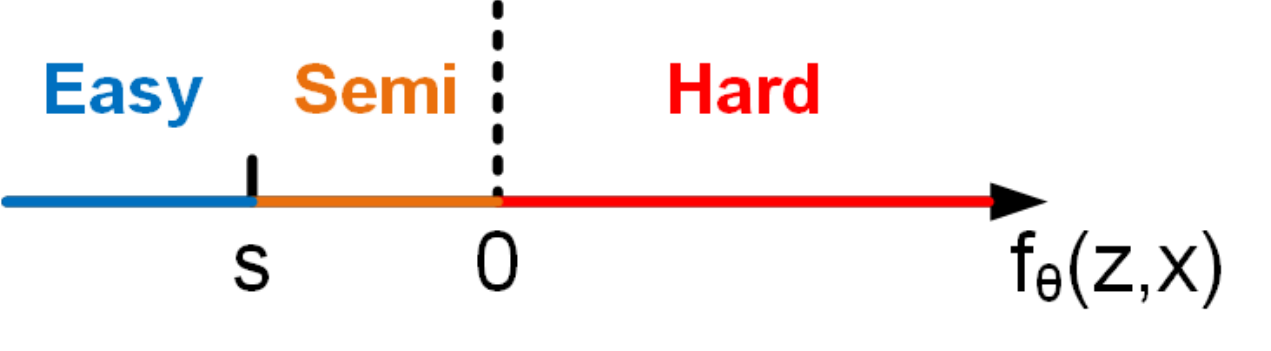}}
\caption{Examples of hard, semi-hard and easy samples for (a) $s>0$, and (b) $s<0$. In the figure, $s$ is the ground truth SDF, and we define the difficulty of each sample according to its estimation $f_\theta(z,x)$.} 
\label{fig:hard}
\end{figure}

Most deep embedding learning methods judge the difficulty of samples according to the loss function~\cite{schroff2015facenet,duan2019deep}. However, the $L_1$-norm loss can be very small for the points with wrong sign estimations. As signs play an important role in implicit representations, we directly define the hard and semi-hard samples based on their sign estimations. More specifically, we consider the points with wrong sign estimations as hard samples, with the estimated SDF values between zero and ground truth values as semi-hard samples, and the others as easy samples. Fig.~\ref{fig:hard} shows the examples. For the semi-hard samples, although currently they obtain correct sign estimations, they are still at high risk of becoming wrong as their predictions are closer to the boundary than the ground truth positions.

To increase the weights of both hard and semi-hard samples, and also decrease the weights of easy samples, we formulate the objective function as below:
\begin{eqnarray} \label{opt_Lepslamp}
L_{\varepsilon,\lambda}(f_\theta(z_i,x_j),s_j) = \left(1 + \lambda sgn(\bar{s}_j)sgn(\bar{s}_j - \bar{f}_\theta(z_i,x_j))\right) L_\varepsilon(f_\theta(z_i,x_j),s_j),
\end{eqnarray}
where $0\leq\lambda<1$ is a hyperparameter controlling the importance of the hard and semi-hard samples, $sgn(v) = 1$ if $v\geq0$ and -1 otherwise.

The physical meaning of (\ref{opt_Lepslamp}) is that we increase the weights of hard and semi-hard samples to $1+\lambda$, and also decrease the weights of easy samples to $1-\lambda$. Although we treat hard and semi-hard samples similarly, their properties are different due to the varying physical meanings as we will demonstrate in the experiments. 
Our hard sample mining strategy always targets at the weakness of the current network rather than using the predefined weights. Still, (\ref{opt_Lepslamp}) will degenerate to (\ref{opt_Leps}) if we set $\lambda=0$. Another understanding of (\ref{opt_Lepslamp}) is that $sgn(\bar{s}_j)$ shows the ground truth sign while $sgn(\bar{s}_j - \bar{f}_\theta(z_i,x_j))$ indicates the direction of optimization. We increase the weights if this direction matches the ground truth sign and decrease the weights otherwise. 

We also design a curriculum for sample difficulty by controlling $\lambda$ at different training stages. At the beginning of training, we aim to teach the network global structures and allow small errors in shape geometry. To this end, we set a relatively small $\lambda$ to make the network equally focused on all training samples. Then, we gradually increase $\lambda$ to emphasize more on hard and semi-hard samples, which helps the network to address its weaknesses and reconstruct better local details. Strictly speaking, the curriculum of sample difficulty is slightly different from the formulation in~\cite{bengio2009curriculum}, as it starts from the original task and gradually increases the difficulty to a harder objective. However, they share similar thoughts and the ablation study also shows the effectiveness of the designed curriculum.

\subsection{Implementation Details.} 
In order to make fair comparisons, we applied the same training data, training epochs and network architecture as DeepSDF~\cite{park2019deepsdf}. More specifically, we prepared the input samples $X_i$ from each shape mesh which was normalized to a unit sphere. We sampled 500,000 points from each shape. The points were sampled more aggressively near the surface to capture more shape details. The learning rate for training the network was set as $N_b\times10^{-5}$ where $N_b$ is the batch size and $10^{-3}$ for the latent vectors. We trained the models for 2,000 epochs. Table~\ref{tab:train} presents the training details, which will degenerate to DeepSDF if we train all the 8 fully connected layers by setting $\varepsilon=\lambda=0$ from beginning to the end. 

\begin{table}[tb]
\centering
\caption{The training details of our method. \emph{Layer} shows the number of fully connected layers. \emph{Residual} represents whether we use a residual block to add layers smoothly.} 
\label{tab:train}
\begin{tabular}{lccccccc} \toprule
Epoch & ~0-200~ & ~200-400~ & ~400-600~ & ~600-800~ & ~800-1000~ & ~1000-1200~ & ~1200-2000~\\ \midrule
Layer & 5 & 6 & 6 & 7 & 7 & 8 & 8 \\
Residual~ & $\times$ & \checkmark & $\times$ & \checkmark & $\times$ & \checkmark & $\times$ \\
$\varepsilon$ & 0.025 & 0.01 & 0.01 & 0.0025 & 0.0025 & 0 & 0 \\
$\lambda$ & 0 & 0.1 & 0.1 & 0.2 & 0.2 & 0.5 & 0.5 \\
 \bottomrule
\end{tabular}
\end{table}

\section{Experiments}
In this section, we perform a thorough comparison of our proposed Curriculum DeepSDF to DeepSDF along with comprehensive ablation studies for the shape reconstruction task on the ShapeNet dataset~\cite{chang2015shapenet}. We use the missing part recovery task as an application to demonstrate the usage of our method.

Following~\cite{park2019deepsdf}, we report the standard distance metrics of mesh reconstruction including the mean and the median of Chamfer distance (CD), mean Earth Mover's distance (EMD)~\cite{rubner2000earth}, and mean mesh accuracy~\cite{seitz2006comparison}. For evaluating CD, we sample 30,000 points from mesh surfaces.
For evaluating EMD, we follow~\cite{park2019deepsdf} by sampling 500 points from mesh surfaces due to a high computation cost.
For evaluating mesh accuracy, following~\cite{seitz2006comparison,park2019deepsdf}, we sample 1,000 points from mesh surfaces and compute the minimum distance $d$ such that 90\% of the points lie within $d$ of the ground truth surface. 

\subsection{Shape Reconstruction}
We conducted experiments on the ShapeNet dataset~\cite{chang2015shapenet} for the shape reconstruction task. In the following, we will introduce quantitative results, ablation studies and visualization results.

\textbf{Quantitative results.} We compare our method to the state-of-the-art methods, including AtlasNet~\cite{groueix2018papier} and DeepSDF~\cite{park2019deepsdf} in Table~\ref{tab:rec}. We also include several variants of our own method for ablation studies. 
\emph{Ours}, representing the proposed Curriculum DeepSDF method, performs a complete curriculum learning considering both surface accuracy and sample difficulty. As variants of our method, \emph{ours-sur} and \emph{ours-sur w/o} only employ the surface accuracy based curriculum learning with/without progressively growth of the network layers, where \emph{ours-sur w/o} uses the fixed architecture with the deepest size; \emph{ours-sam} only employs sample difficulty based curriculum learning. For a fair comparison, we evaluated all SDF-based methods following the same training and testing protocols as DeepSDF, including training/test split, the number of training epochs, and network architecture, etc. For AtlasNet-based methods, we directly report the numbers from~\cite{park2019deepsdf}. Here are the three key observations from Table~\ref{tab:rec}:
\begin{enumerate}
    \item Compared to vanilla DeepSDF, curriculum learning on either surface accuracy or sample difficulty can lead to a significant performance gain. The best performance is achieved by simultaneously performing both curricula. 
    
    \item In general, the curriculum of sample difficulty helps more on lamp and plane as these categories suffer more from reconstructing slender or thin structures. The curriculum of surface accuracy is more effective for the categories of chair, sofa and table where shapes are more regular.
    
    
    \item As we only sample 500 points for computing EMD, even the ground truth mesh has non-zero EMD to itself rising from the randomness in point sampling. 
    Our performance is approaching the upper bound on plane and sofa. 
\end{enumerate}

\begin{table}[!htb]
\centering
\caption{\titlecap{Reconstructing shapes from the ShapeNet test set.} Here we report shape reconstruction errors in term of several distance metrics on five ShapeNet classes. Note that we multiply CD by $10^3$ and mesh accuracy by $10^1$. The \emph{average} column shows the average distance and the \emph{relative} column shows the relative distance reduction compared to DeepSDF. For all metrics except for \emph{relative}, the lower, the better.} 
\label{tab:rec}
\begin{tabular}{lp{1.3cm}<{\centering}p{1.3cm}<{\centering}p{1.3cm}<{\centering}p{1.3cm}<{\centering}p{1.3cm}<{\centering}p{1.3cm}<{\centering}p{1.3cm}<{\centering}p{1.3cm}<{\centering}} \toprule
CD, mean & lamp & plane & chair & sofa & table & average & relative\\ \midrule
AtlasNet-Sph & 2.381 & 0.188 & 0.752 & 0.445 & 0.725 & 0.730 & - \\
AtlasNet-25 & 1.182 & 0.216 & 0.368 & 0.411 & 0.328 & 0.391 & - \\
DeepSDF & 0.776 & 0.143 & 0.243 & 0.117 & 0.424 & 0.319 & - \\
Ours-Sur w/o & 0.743 & 0.109 & 0.162 & 0.110 & 0.343 & 0.257 & 19.4\% \\
Ours-Sur & 0.639 & 0.086 & 0.157 & 0.108 & 0.327 & 0.239 & 25.1\% \\
Ours-Sam & 0.592 & 0.078 & 0.175 & 0.113 & 0.342 & 0.246 & 22.9\% \\
Ours & \textbf{0.473} & \textbf{0.070} & \textbf{0.156} & \textbf{0.105} & \textbf{0.304} & \textbf{0.216} & \textbf{32.3\%} \\ \midrule
CD, median & & & & & & & \\ \midrule
AtlasNet-Sph & 2.180 & 0.079 & 0.511 & 0.330 & 0.389 & 0.490 & - \\
AtlasNet-25 & 0.993 & 0.065 & 0.276 & 0.311 & 0.195 & 0.267 & - \\
DeepSDF & 0.178 & 0.061 & 0.098 & 0.081 & 0.052 & 0.078 & - \\
Ours-Sur w/o & 0.172 & 0.048 & 0.071 & 0.077 & 0.047 & 0.066 & 15.4\% \\
Ours-Sur & 0.147 & 0.045 & \textbf{0.064} & 0.077 & 0.051 & 0.063 & 19.2\% \\
Ours-Sam & 0.139 & 0.040 & 0.066 & 0.080 & 0.050 & 0.063 & 19.2\% \\
Ours & \textbf{0.105} & \textbf{0.033} & \textbf{0.064} & \textbf{0.069} & \textbf{0.048} & \textbf{0.056} & \textbf{28.2\%} \\ \midrule
EMD, mean & & & & & & & \\ \midrule
GT & 0.034 & 0.026 & 0.041 & 0.044 & 0.041 & 0.039 & - \\
AtlasNet-Sph & 0.085 & 0.038 & 0.071 & 0.050 & 0.060 & 0.060 & - \\
AtlasNet-25 & 0.062 & 0.041 & 0.064 & 0.063 & 0.073 & 0.064 & - \\
DeepSDF & 0.066 & 0.035 & 0.055 & 0.051 & 0.057 & 0.053 & - \\
Ours-Sur w/o & 0.057 & 0.032 & \textbf{0.048} & 0.046 & 0.049 & 0.046 & 13.2\% \\
Ours-Sur & 0.055 & 0.027 & \textbf{0.048} & 0.046 & \textbf{0.048} & 0.045 & 15.1\% \\
Ours-Sam & 0.055 & 0.027 & 0.053 & 0.050 & 0.051 & 0.048 & 9.4\% \\
Ours & \textbf{0.052} & \textbf{0.026} & \textbf{0.048} & \textbf{0.044} & \textbf{0.048} & \textbf{0.044} & \textbf{17.0\%} \\ \midrule
Mesh acc, mean & & & & & & & \\ \midrule
AtlasNet-Sph & 0.540 & 0.130 & 0.330 & 0.170 & 0.320 & 0.290 & - \\
AtlasNet-25 & 0.420 & 0.130 & 0.180 & 0.170 & 0.140 & 0.172 & - \\
DeepSDF & 0.155 & 0.044 & 0.104 & 0.041 & 0.120 & 0.097 & - \\
Ours-Sur w/o & 0.133 & 0.035 & 0.089 & 0.040 & 0.104 & 0.083 & 14.4\% \\
Ours-Sur & 0.121 & 0.034 & 0.082 & 0.039 & 0.098 & 0.078 & 19.6\% \\
Ours-Sam & 0.135 & \textbf{0.031} & 0.083 & \textbf{0.036} & 0.087 & 0.074 & 23.7\% \\
Ours & \textbf{0.103} & \textbf{0.031} & \textbf{0.080} & \textbf{0.036} & \textbf{0.087} & \textbf{0.071} & \textbf{26.8\%} \\
 \bottomrule
\end{tabular}
\end{table} 


\textbf{Hard sample mining strategies.} 
We conducted ablation studies for a more detailed analysis of different hard sample mining strategies
on the lamp category due to its large variations and complex shape details. In the curriculum of sample difficulty, we gradually increase $\lambda$ to make the network more and more focused on the hard samples. We compared it with the simple strategy by fixing a single $\lambda$. Table~\ref{tab:lam} shows that the performance improves as $\lambda$ increases until reaching a sweet spot, after which further increasing $\lambda$ could hurt the performance.
The best result is achieved by our method which gradually increases $\lambda$ as it encourages the network to focus more and more on hard details. 

\begin{table}[tb]
\centering
\caption{Experimental comparisons with using fixed $\lambda$ for hard sample mining. The method degenerates to \emph{ours-sur} when $\lambda=0$. CD is multiplied by $10^3$.} 
\label{tab:lam}
\begin{tabular}{p{1.5cm}p{1.3cm}<{\centering}p{1.3cm}<{\centering}p{1.3cm}<{\centering}p{1.3cm}<{\centering}p{1.3cm}<{\centering}p{1.3cm}<{\centering}p{1.3cm}<{\centering}} \toprule
$\lambda$ & 0 & 0.05 & 0.10 & 0.25 & 0.50 & 0.75 & Ours \\ \midrule
CD, mean & 0.639 & 0.606 & 0.549 & 0.538 & 0.508 & 0.567 & \textbf{0.473} \\
 \bottomrule
\end{tabular}
\end{table}

\begin{table}[tb]
\centering
\caption{Experimental comparisons of different hard sample mining strategies. In the table, \emph{H}, \emph{S} and \emph{E} are the hard, semi-hard and easy samples, respectively. For the symbols, $\uparrow$ is to increase the weights to $1+\lambda$, $\downarrow$ is to decrease the weights to $1-\lambda$ and - is to maintain the weights. \emph{H($\uparrow$)S($\uparrow$)E($\downarrow$)} is the sampling strategy used in our method, while \emph{H(-)S(-)E(-)} degenerates to \emph{ours-sur}. CD is multiplied by $10^3$.} 
\label{tab:hard}
\begin{tabular}{p{1.6cm}p{2.2cm}<{\centering}p{2.2cm}<{\centering}p{2.2cm}<{\centering}p{2.2cm}<{\centering}} \toprule
Strategy & H(-)S(-)E(-) & H(-)S(-)E($\downarrow$) & H(-)S($\uparrow$)E(-) & H(-)S($\uparrow$)E($\downarrow$) \\ \midrule
CD, mean & 0.639 & 0.563 & 0.587 & 0.508 \\ \midrule \midrule
Strategy & H($\uparrow$)S(-)E(-) & H($\uparrow$)S(-)E($\downarrow$) & H($\uparrow$)S($\uparrow$)E(-) & H($\uparrow$)S($\uparrow$)E($\downarrow$) \\ \midrule
CD, mean & 0.676 & 0.661 & 0.512 & \textbf{0.473} \\
 \bottomrule
\end{tabular}
\end{table}

For hard sample mining,
we increase the weights of hard and semi-hard samples to $1+\lambda$ and also decrease the weights of easy samples to $1-\lambda$
. As various similar strategies can be used, we 
demonstrate the effectiveness of our design in Table~\ref{tab:hard}
. We observe that both increasing the weights of semi-hard samples and decreasing the weights of easy samples can boost the performance. However, it is risky to only increase weights for hard samples excluding semi-hard ones in which case the performance drops. One possible reason is that focusing too much on hard samples may lead to more wrong sign estimations for the semi-hard ones as they are close to the boundary. Hence, it is necessary to increase the weights of semi-hard samples as well to maintain their correct sign estimations. The best performance is achieved by simultaneously increasing the weights of hard and semi-hard samples and decreasing the weights of easy ones.

\begin{table}[tb]
\centering
\caption{Comparison of mean of EMD on the lamp category of the ShapeNet dataset with varying numbers of sampled points.} 
\label{tab:emd}
\begin{tabular}{p{2.5cm}p{1.5cm}<{\centering}p{1.5cm}<{\centering}p{1.5cm}<{\centering}p{1.5cm}<{\centering}p{1.5cm}<{\centering}} \toprule
Number of points & 500 & 2000 & 5000 & 10000 \\ \midrule
GT & 0.034 & 0.008 & 0.008 & 0.004 \\
DeepSDF & 0.066 & 0.056 & 0.052 & 0.051 \\
Ours-Sur w/o & 0.057 & 0.053 & 0.050 & 0.048 \\
Ours-Sur & 0.055 & 0.052 & 0.049 & 0.048 \\
Ours-Sam & 0.055 & 0.053 & 0.050 & 0.048 \\
Ours & \textbf{0.052} & \textbf{0.051} & \textbf{0.047} & \textbf{0.046} \\
 \bottomrule
\end{tabular}
\end{table}

\textbf{Number of points for EMD.} In Table~\ref{tab:rec}, we followed~\cite{park2019deepsdf} by sampling 500 points to compute accurate EMD, which would lead to relatively large distance even for 
ground truth meshes. To this end, we increase the number of sampled points during EMD computation and tested the performance on lamps. Results in Table~\ref{tab:emd} show
that the number of sampled points can affect EMD due to the randomness in sampling, and the EMD of resampled ground truth decreases when using more points. Our method continuously obtains better results. 

\begin{figure}[tb]
\centering
\includegraphics[width=0.96\textwidth]{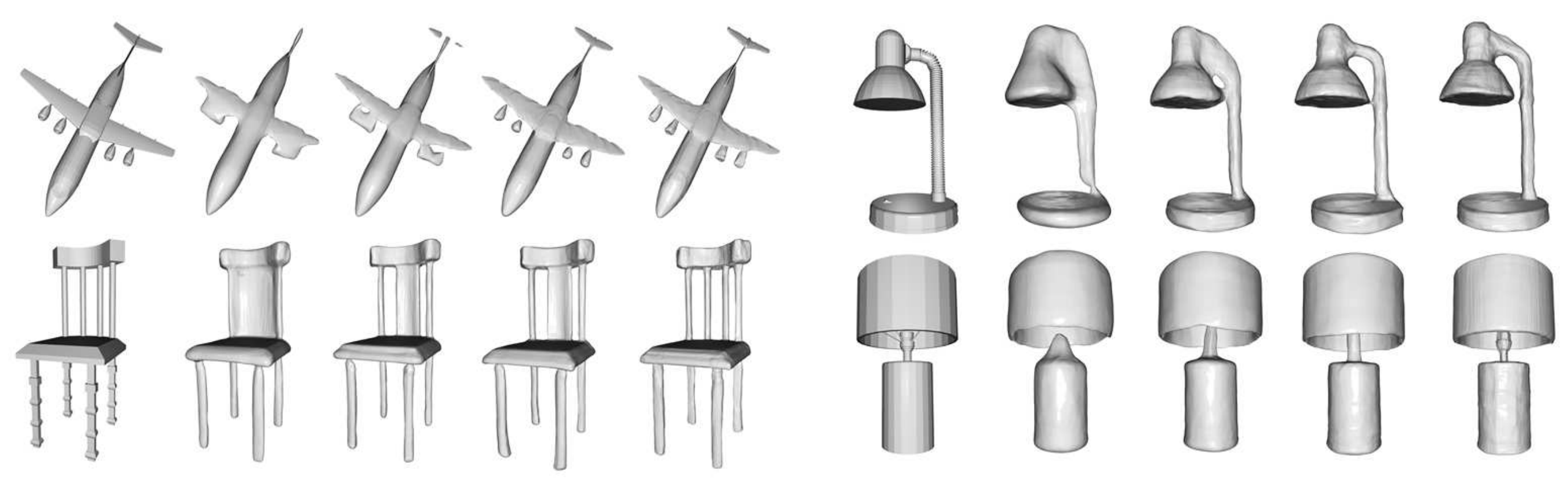}
\caption{The visualization of shape reconstruction at the end of each training stage. From left to right: ground truth, 200 epochs, 600 epochs, 1000 epochs, and 2000 epochs.}
\label{fig:epoch}
\end{figure}

\textbf{Visualization results.} We visualize the shape reconstruction results in Fig.~\ref{fig:vis} to qualitatively compare DeepSDF and Curriculum DeepSDF.
We observe that Curriculum DeepSDF reconstructs more accurate shape surfaces. The curriculum of surface accuracy helps to better capture the general structure, and sample difficulty encourages the recovery of complex local details. We also provide the reconstructed shapes at key epochs in Fig.~\ref{fig:epoch}. Curriculum DeepSDF learns coarse shapes at early stages which omits complex details. Then, it gradually refines local parts based on the learned coarse shapes. This training procedure improves the performance of the learned shape representation.

\begin{table}[tb]
\centering
\caption{Experimental comparisons under different ratios of removed points. CD and mesh accuracy are multiplied by $10^3$ and $10^1$, respectively.} 
\label{tab:part}
\begin{tabular}{p{1.42cm}p{0.9cm}<{\centering}p{0.9cm}<{\centering}p{0.08cm}<{\centering}p{0.9cm}<{\centering}p{0.9cm}<{\centering}p{0.08cm}<{\centering}p{0.9cm}<{\centering}p{0.9cm}<{\centering}p{0.08cm}<{\centering}p{0.9cm}<{\centering}p{0.9cm}<{\centering}p{0.08cm}<{\centering}p{0.9cm}<{\centering}p{0.9cm}<{\centering}} \toprule
Method & \multicolumn{2}{c}{5\%} && \multicolumn{2}{c}{10\%} && \multicolumn{2}{c}{15\%} && \multicolumn{2}{c}{20\%} && \multicolumn{2}{c}{25\%} \\
\cline{2-3} \cline{5-6} \cline{8-9} \cline{11-12} \cline{14-15} \\ [-8pt]
$\backslash$Metric & CD & Mesh && CD & Mesh && CD & Mesh && CD & Mesh && CD & Mesh \\ \midrule \\ [-13pt]
plane & & && & && & && & && & \\ [-2pt] \midrule
DeepSDF & 0.163 & 0.056 && 0.229 & 0.066 && 0.217 & 0.067 && 0.224 & 0.069 && 0.233 & 0.080 \\
Ours & \textbf{0.095} & \textbf{0.032} && \textbf{0.124} & \textbf{0.044} && \textbf{0.149} & \textbf{0.052} && \textbf{0.163} & \textbf{0.062} && \textbf{0.192} & \textbf{0.072} \\ \midrule \\ [-13pt]
sofa & & && & && & && & && & \\ [-2pt] \midrule
DeepSDF & 0.133 & 0.045 && 0.137 & 0.047 && 0.149 & 0.050 && 0.169 & 0.058 && \textbf{0.196} & 0.066 \\
Ours & \textbf{0.110} & \textbf{0.037} && \textbf{0.120} & \textbf{0.041} && \textbf{0.143} & \textbf{0.046} && \textbf{0.165} & \textbf{0.053} && \textbf{0.196} & \textbf{0.061} \\ \midrule \\ [-13pt]
lamp & & && & && & && & && & \\ [-2pt] \midrule
DeepSDF & 2.08 & 0.230 && 3.10 & 0.241 && 3.50 & 0.286 && 4.18 & 0.307 && 4.79 & 0.331 \\
Ours & \textbf{1.96} & \textbf{0.167} && \textbf{2.87} & \textbf{0.195} && \textbf{3.27} & \textbf{0.231} && \textbf{3.52} & \textbf{0.277} && \textbf{4.07} & \textbf{0.320} \\ 
 \bottomrule
\end{tabular}
\end{table}

\subsection{Missing Part Recovery}
One of the main advantages of the DeepSDF framework is that we can optimize a shape code based upon a partial shape observation, and then render the complete shape through the learned network. In this subsection, we compare DeepSDF with Curriculum DeepSDF on the task of missing part recovery.

\begin{figure}[tb]
\centering
\includegraphics[width=0.88\textwidth]{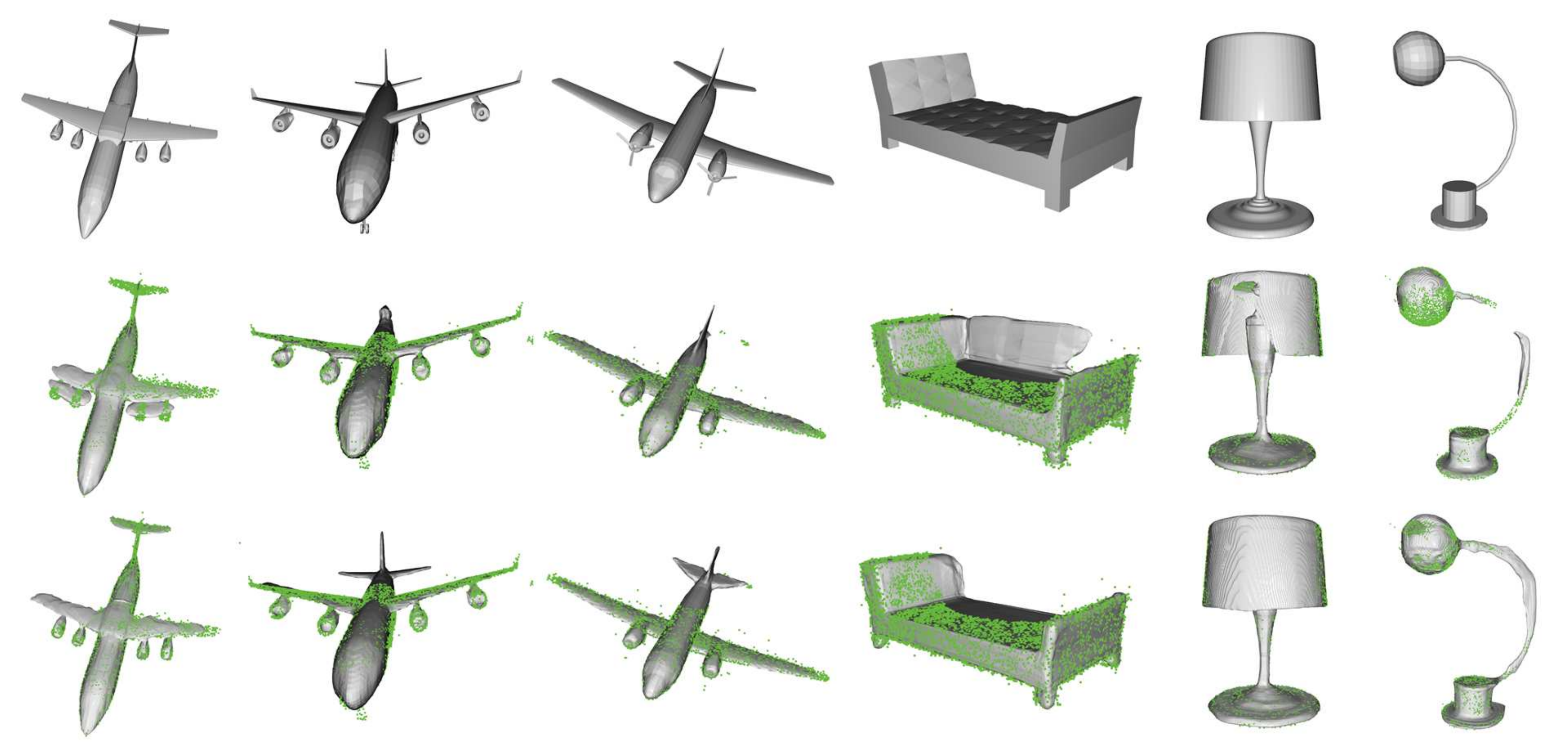}
\caption{The visualization results of missing part recovery. The green points are the remaining points that we use to recover the whole mesh. From top to bottom: ground truth, DeepSDF, and Curriculum DeepSDF.}
\label{fig:recovery}
\end{figure}

To create partial shapes with missing parts, we remove a subset of points from each shape $X_i$
As random point removal may still preserve the holistic structures, we remove all the points in a local area to create missing parts. More specifically, we randomly select a point from the shape and then remove a certain quantity of its nearest neighbor points including itself, so that all the points within a local range can be removed. We conducted the experiments on three ShapeNet categories: plane, sofa and lamp. In these categories, plane and sofa have more regular and symmetric structures, while lamp is more complex and contains large variations. Table~\ref{tab:part} shows 
that part removal largely affects the performance on the lamp category compared with plane and sofa, and Curriculum DeepSDF continuously obtains better results than DeepSDF under different ratios of removed points. A visual comparison is provided in Fig.~\ref{fig:recovery}.

\section{Conclusion}
In this paper, we have proposed Curriculum DeepSDF by designing a shape curriculum for shape representation learning. Inspired by the learning principle of humans, we organize the learning task into a series of difficulty levels from surface accuracy and sample difficulty. For surface accuracy, we design a tolerance parameter to control the global smoothness, which 
gradually increases the accuracy of the learned shape with more layers. For sample difficulty, we define hard, semi-hard and easy training samples in SDF learning, 
and gradually re-weight the samples to 
focus more and more on difficult local details. Experimental results show that our method largely improves the performance of DeepSDF with the same training data, training epochs and network architecture.

\section*{Acknowledgements}
This research was supported by a Vannevar Bush Faculty Fellowship, a grant from the SAIL-Toyota Center for AI Research, and gifts from Amazon Web Services and the Dassault Foundation.

\bibliographystyle{splncs04}
\bibliography{egbib}

\begin{thebibliography}{10}
\providecommand{\url}[1]{\texttt{#1}}
\providecommand{\urlprefix}{URL }
\providecommand{\doi}[1]{https://doi.org/#1}

\bibitem{achlioptas2018learning}
Achlioptas, P., Diamanti, O., Mitliagkas, I., Guibas, L.: Learning
  representations and generative models for {3D} point clouds. In: ICML. pp.
  40--49 (2018)

\bibitem{allgower2003introduction}
Allgower, E.L., Georg, K.: Introduction to numerical continuation methods,
  vol.~45. SIAM (2003)

\bibitem{bengio2009curriculum}
Bengio, Y., Louradour, J., Collobert, R., Weston, J.: Curriculum learning. In:
  ICML. pp. 41--48 (2009)

\bibitem{carr2001reconstruction}
Carr, J.C., Beatson, R.K., Cherrie, J.B., Mitchell, T.J., Fright, W.R.,
  McCallum, B.C., Evans, T.R.: Reconstruction and representation of {3D}
  objects with radial basis functions. In: SIGGRAPH. pp. 67--76 (2001)

\bibitem{chang2015shapenet}
Chang, A.X., Funkhouser, T., Guibas, L., Hanrahan, P., Huang, Q., Li, Z.,
  Savarese, S., Savva, M., Song, S., Su, H., et~al.: {ShapeNet}: An
  information-rich {3D} model repository. arXiv preprint arXiv:1512.03012
  (2015)

\bibitem{chen2019learning}
Chen, Z., Zhang, H.: Learning implicit fields for generative shape modeling.
  In: CVPR. pp. 5939--5948 (2019)

\bibitem{choy20163d}
Choy, C.B., Xu, D., Gwak, J., Chen, K., Savarese, S.: {3D-R2N2}: A unified
  approach for single and multi-view 3d object reconstruction. In: ECCV. pp.
  628--644 (2016)

\bibitem{duan2019deep}
Duan, Y., Chen, L., Lu, J., Zhou, J.: Deep embedding learning with
  discriminative sampling policy. In: CVPR. pp. 4964--4973 (2019)

\bibitem{duan2020deep}
Duan, Y., Lu, J., Zheng, W., Zhou, J.: Deep adversarial metric learning. TIP
  \textbf{29}(1),  2037--2051 (2020)

\bibitem{duan2019structural}
Duan, Y., Zheng, Y., Lu, J., Zhou, J., Tian, Q.: Structural relational
  reasoning of point clouds. In: CVPR. pp. 949--958 (2019)

\bibitem{elman1993learning}
Elman, J.L.: Learning and development in neural networks: The importance of
  starting small. Cognition  \textbf{48}(1),  71--99 (1993)

\bibitem{genova2019deep}
Genova, K., Cole, F., Sud, A., Sarna, A., Funkhouser, T.: Deep structured
  implicit functions. arXiv preprint arXiv:1912.06126  (2019)

\bibitem{genova2019learning}
Genova, K., Cole, F., Vlasic, D., Sarna, A., Freeman, W.T., Funkhouser, T.:
  Learning shape templates with structured implicit functions. In: ICCV. pp.
  7154--7164 (2019)

\bibitem{graves2017automated}
Graves, A., Bellemare, M.G., Menick, J., Munos, R., Kavukcuoglu, K.: Automated
  curriculum learning for neural networks. In: ICML. pp. 1311--1320 (2017)

\bibitem{gropp2020implicit}
Gropp, A., Yariv, L., Haim, N., Atzmon, M., Lipman, Y.: Implicit geometric
  regularization for learning shapes. arXiv preprint arXiv:2002.10099  (2020)

\bibitem{groueix2018papier}
Groueix, T., Fisher, M., Kim, V.G., Russell, B.C., Aubry, M.: A
  papier-m{\^a}ch{\'e} approach to learning {3D} surface generation. In: CVPR.
  pp. 216--224 (2018)

\bibitem{guo20153d}
Guo, K., Zou, D., Chen, X.: {3D} mesh labeling via deep convolutional neural
  networks. TOG  \textbf{35}(1),  1--12 (2015)

\bibitem{hacohen2019power}
Hacohen, G., Weinshall, D.: On the power of curriculum learning in training
  deep networks. In: ICML. pp. 2535--2544 (2019)

\bibitem{ilg2017flownet}
Ilg, E., Mayer, N., Saikia, T., Keuper, M., Dosovitskiy, A., Brox, T.: Flownet
  2.0: Evolution of optical flow estimation with deep networks. In: CVPR. pp.
  2462--2470 (2017)

\bibitem{jiang2018mentornet}
Jiang, L., Zhou, Z., Leung, T., Li, L.J., Fei-Fei, L.: {MentorNet}: Learning
  data-driven curriculum for very deep neural networks on corrupted labels. In:
  ICML. pp. 2304--2313 (2018)

\bibitem{karras2017progressive}
Karras, T., Aila, T., Laine, S., Lehtinen, J.: Progressive growing of gans for
  improved quality, stability, and variation. arXiv preprint arXiv:1710.10196
  (2017)

\bibitem{liao2018deep}
Liao, Y., Donne, S., Geiger, A.: Deep marching cubes: Learning explicit surface
  representations. In: CVPR. pp. 2916--2925 (2018)

\bibitem{liu2019learning}
Liu, S., Saito, S., Chen, W., Li, H.: Learning to infer implicit surfaces
  without {3D} supervision. In: NeurIPS. pp. 8293--8304 (2019)

\bibitem{lorensen1987marching}
Lorensen, W.E., Cline, H.E.: Marching cubes: A high resolution {3D} surface
  construction algorithm. SIGGRAPH pp. 163--169 (1987)

\bibitem{lu2017discriminative}
Lu, J., Hu, J., Tan, Y.P.: Discriminative deep metric learning for face and
  kinship verification. TIP  \textbf{26}(9),  4269--4282 (2017)

\bibitem{maturana2015voxnet}
Maturana, D., Scherer, S.: Voxnet: A {3D} convolutional neural network for
  real-time object recognition. In: IROS. pp. 922--928 (2015)

\bibitem{mescheder2019occupancy}
Mescheder, L., Oechsle, M., Niemeyer, M., Nowozin, S., Geiger, A.: Occupancy
  networks: Learning {3D} reconstruction in function space. In: CVPR. pp.
  4460--4470 (2019)

\bibitem{michalkiewicz2019deep}
Michalkiewicz, M., Pontes, J.K., Jack, D., Baktashmotlagh, M., Eriksson, A.:
  Deep level sets: Implicit surface representations for {3D} shape inference.
  arXiv preprint arXiv:1901.06802  (2019)

\bibitem{ohtake2005multi}
Ohtake, Y., Belyaev, A., Alexa, M., Turk, G., Seidel, H.P.: Multi-level
  partition of unity implicits. In: SIGGRAPH. pp. 173--180 (2005)

\bibitem{park2019deepsdf}
Park, J.J., Florence, P., Straub, J., Newcombe, R., Lovegrove, S.: {DeepSDF}:
  Learning continuous signed distance functions for shape representation. In:
  CVPR. pp. 165--174 (2019)

\bibitem{qi2017pointnet}
Qi, C.R., Su, H., Mo, K., Guibas, L.J.: Pointnet: Deep learning on point sets
  for {3D} classification and segmentation. In: CVPR. pp. 652--660 (2017)

\bibitem{qi2016volumetric}
Qi, C.R., Su, H., Nie{\ss}ner, M., Dai, A., Yan, M., Guibas, L.J.: Volumetric
  and multi-view cnns for object classification on {3D} data. In: CVPR. pp.
  5648--5656 (2016)

\bibitem{qi2017pointnet++}
Qi, C.R., Yi, L., Su, H., Guibas, L.J.: Pointnet++: Deep hierarchical feature
  learning on point sets in a metric space. In: NeurIPS. pp. 5099--5108 (2017)

\bibitem{rao2020global}
Rao, Y., Lu, J., Zhou, J.: Global-local bidirectional reasoning for
  unsupervised representation learning of {3D} point clouds. In: CVPR. pp.
  5376--5385 (2020)

\bibitem{rubner2000earth}
Rubner, Y., Tomasi, C., Guibas, L.J.: The earth mover's distance as a metric
  for image retrieval. IJCV  \textbf{40}(2),  99--121 (2000)

\bibitem{saito2019pifu}
Saito, S., Huang, Z., Natsume, R., Morishima, S., Kanazawa, A., Li, H.: {PIFu}:
  Pixel-aligned implicit function for high-resolution clothed human
  digitization. In: ICCV. pp. 2304--2314 (2019)

\bibitem{sanger1994neural}
Sanger, T.D.: Neural network learning control of robot manipulators using
  gradually increasing task difficulty. IEEE transactions on Robotics and
  Automation  \textbf{10}(3),  323--333 (1994)

\bibitem{schroff2015facenet}
Schroff, F., Kalenichenko, D., Philbin, J.: {FaceNet}: A unified embedding for
  face recognition and clustering. In: CVPR. pp. 815--823 (2015)

\bibitem{seitz2006comparison}
Seitz, S.M., Curless, B., Diebel, J., Scharstein, D., Szeliski, R.: A
  comparison and evaluation of multi-view stereo reconstruction algorithms. In:
  CVPR. pp. 519--528 (2006)

\bibitem{sharma2018improved}
Sharma, R., Barratt, S., Ermon, S., Pande, V.: Improved training with
  curriculum {GANs}. arXiv preprint arXiv:1807.09295  (2018)

\bibitem{shen2004interpolating}
Shen, C., O'Brien, J.F., Shewchuk, J.R.: Interpolating and approximating
  implicit surfaces from polygon soup. In: SIGGRAPH. pp. 896--904 (2004)

\bibitem{sinha2016deep}
Sinha, A., Bai, J., Ramani, K.: Deep learning {3D} shape surfaces using
  geometry images. In: ECCV. pp. 223--240 (2016)

\bibitem{su2015multi}
Su, H., Maji, S., Kalogerakis, E., Learned-Miller, E.: Multi-view convolutional
  neural networks for {3D} shape recognition. In: ICCV. pp. 945--953 (2015)

\bibitem{sun2020circle}
Sun, Y., Cheng, C., Zhang, Y., Zhang, C., Zheng, L., Wang, Z., Wei, Y.: Circle
  loss: A unified perspective of pair similarity optimization. In: CVPR. pp.
  6398--6407 (2020)

\bibitem{tulsiani2017multi}
Tulsiani, S., Zhou, T., Efros, A.A., Malik, J.: Multi-view supervision for
  single-view reconstruction via differentiable ray consistency. In: CVPR. pp.
  2626--2634 (2017)

\bibitem{turk1999shape}
Turk, G., O'brien, J.F.: Shape transformation using variational implicit
  functions. In: SIGGRAPH. pp. 14--20 (1999)

\bibitem{turk2002modelling}
Turk, G., O'brien, J.F.: Modelling with implicit surfaces that interpolate. TOG
   \textbf{21}(4),  855--873 (2002)

\bibitem{wang2018pixel2mesh}
Wang, N., Zhang, Y., Li, Z., Fu, Y., Liu, W., Jiang, Y.G.: Pixel2mesh:
  Generating {3D} mesh models from single {RGB} images. In: ECCV. pp. 52--67
  (2018)

\bibitem{wang2019dynamic}
Wang, Y., Sun, Y., Liu, Z., Sarma, S.E., Bronstein, M.M., Solomon, J.M.:
  Dynamic graph {CNN} for learning on point clouds. TOG)  \textbf{38}(5),
  1--12 (2019)

\bibitem{weinshall2018curriculum}
Weinshall, D., Cohen, G., Amir, D.: Curriculum learning by transfer learning:
  Theory and experiments with deep networks. In: ICML. pp. 5238--5246 (2018)

\bibitem{wu20153d}
Wu, Z., Song, S., Khosla, A., Yu, F., Zhang, L., Tang, X., Xiao, J.: {3D
  ShapeNets}: A deep representation for volumetric shapes. In: CVPR. pp.
  1912--1920 (2015)

\bibitem{xu2019disn}
Xu, Q., Wang, W., Ceylan, D., Mech, R., Neumann, U.: {DISN}: Deep implicit
  surface network for high-quality single-view {3D} reconstruction. In:
  NeurIPS. pp. 490--500 (2019)

\end{thebibliography}
\end{document}